\documentclass[10pt,twocolumn,letterpaper]{article}

\usepackage{cvpr}              

\usepackage{graphicx}
\usepackage{amsmath}
\usepackage{amssymb}
\usepackage{booktabs}
\usepackage{tabularx}
\usepackage{comment}
\usepackage{fontawesome}
\usepackage{cuted}

\usepackage[pagebackref,breaklinks,colorlinks]{hyperref}

\usepackage[capitalize]{cleveref}
\crefname{section}{Sec.}{Secs.}
\Crefname{section}{Section}{Sections}
\Crefname{table}{Table}{Tables}
\crefname{table}{Tab.}{Tabs.}

\def\cvprPaperID{*****} 
\def\confName{CVPR}
\def\confYear{2023}

\begin{document}

\title{Resource Constrained Semantic Segmentation for Waste Sorting}

\author{Cascina Elisa\\
Politecnico di Torino\\
elisa.cascina@studenti.polito.it\\
\and
Pellegrino Andrea\\
Politecnico di Torino\\
s309855@studenti.polito.it\\
\and
Tozzi Lorenzo\\
Politecnico di Torino\\
lorenzo.tozzi@studenti.polito.it\\
}
\maketitle

\begin{abstract}
   This work addresses the need for efficient waste sorting strategies in Materials Recovery Facilities to minimize the environmental impact of rising waste. We propose resource-constrained semantic segmentation models for segmenting recyclable waste in industrial settings. Our goal is to develop models that fit within a 10MB memory constraint, suitable for edge applications with limited processing capacity. We perform the experiments on three networks: ICNet, BiSeNet (Xception39 backbone), and ENet. Given the aforementioned limitation, we implement quantization and pruning techniques on the broader nets, achieving positive results while marginally impacting the Mean IoU metric. Furthermore, we propose a combination of Focal and Lov{\'a}sz loss that addresses the implicit class imbalance resulting in better performance compared with the Cross-entropy loss function.

\end{abstract}

\faGithub \; \textit{\href{https://github.com/anubis09/Resource_Constrained_Semantic_Segmentation_for_Waste_Sorting}{GitHub repository}}

\section{Introduction}
\label{sec:intro}
The exponential growth of the global population has resulted in a significant increase in waste production. To tackle this challenge, it is essential to develop efficient strategies for Materials Recovery Facilities, as they play a vital role in the recycling process. These strategies should focus on improving the detection of recyclable waste and minimizing negative environmental consequences.

\begin{figure}[ht]
    \centering
    \includegraphics[width=0.47\textwidth]{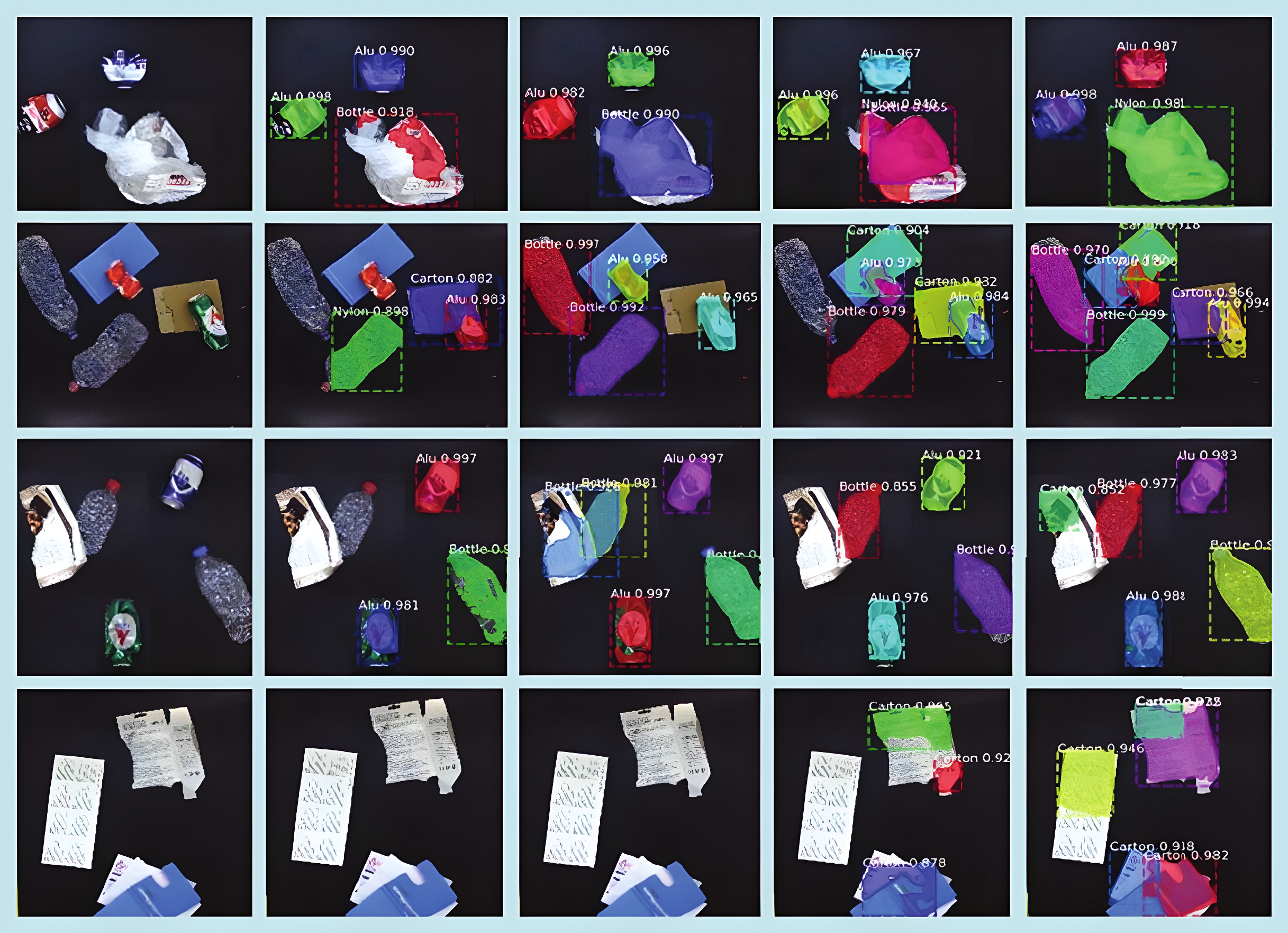}
    \caption{Example of semantic segmentation model result \cite{wasteSorting}}
    \label{fig:dataset with masks}
\end{figure}

Koskinopoulou \etal \cite{wasteSorting} introduced the task of segmenting waste that could be useful for waste sorting. Figure \ref{fig:dataset with masks} shows the qualitative performance by Mask RCNN. However, Mask RCNN models are generally large and have a high number of training parameters. So, to achieve real-time waste segmentation performance, the model should be computationally efficient. 
In this work, we focus on developing resource-constrained semantic segmentation models specifically designed for waste-sorting applications in industrial settings. The primary objective of this work was to deploy tiny semantic segmentation models capable of fitting within a memory constraint of 10MB with significant mIOU. This design makes them suitable for edge applications utilizing smart cameras with limited onboard processing capacity. 

The contribution of this work can be summarized in three folds: Firstly, we select three resource-efficient semantic segmentation network architectures for experimentation, namely: ICNet \cite{icnet}, BiSeNet (with Xception39 backbone) \cite{bisenet,bisenet2} and ENet \cite{enet}. We chose the models above as they show significant performance on semantic segmentation tasks and have great potential for adaptation in resource-constrained environments. Next, we train the models using an augmented dataset with various loss functions to improve performance. We address the class imbalance problem by introducing the models on a plethora of losses: Focal \cite{focal}, Dice \cite{dice}, Class-balanced focal \cite{class-balanced}, as well as the combination of Focal and Lov{\'a}sz\cite{focal+lovasz}. Finally, we ensure parameter efficiency of our model while preserving mIoU results by performing model pruning \cite{pruning_pytorch,pruning_frankle2019lottery} and quantization \cite{quantization_pytorch,quantisation_gholami2021survey,quantisation_krishnamoorthi2018quantizing,quantisation_wu2020integer} techniques. Moreover, given the relatively compact nature of ENet, we opted to augment its dimensions by modifying the sizes of both the encoder and decoder to enhance standard ENet performances while respecting the memory constraint.



\section{Related Work}

Our research builds upon the work by Koskinopoulou \etal \cite{wasteSorting}, which introduces an integrated robotic system designed to sort recyclable materials. This system comprises two principal components: a robotic manipulator for physically segregating waste into distinct bins based on material type and a vision-based module dedicated to material detection and categorization. The materials are Aluminium, Paper, Bottle, and Nylon.

Our study further elaborates on the aspect of vision-based detection.
To accurately categorize recyclables based on their material type in an autonomous recovery system, they chose Convolutional Neural Networks (CNNs), the most cost-effective option.
The study focuses on instance segmentation to identify and label multiple objects within waste images. To accomplish this, the Mask Regional CNN network was employed due to its success in similar tasks. Mask R-CNN offers a scalable method for categorizing recyclables across numerous classes.

The study considers various factors to optimize performances, including tuning training parameters, dataset size, learning steps per epoch, and entation techniques. Multiple versions of the Mask R-CNN model (Net 1 to Net 4) are examined, each with its strengths and limitations. 
Net 1 and Net 2, both trained with smaller datasets, had limitations and poor performance when applied to the waste images. Meanwhile, Net 3 showed significant improvement and could be used in real-world applications. Net 4, which incorporates entation and extended training, outperforms other versions when tested on the final images.

\begin{figure}
    \centering
    \includegraphics[width=0.47\textwidth]{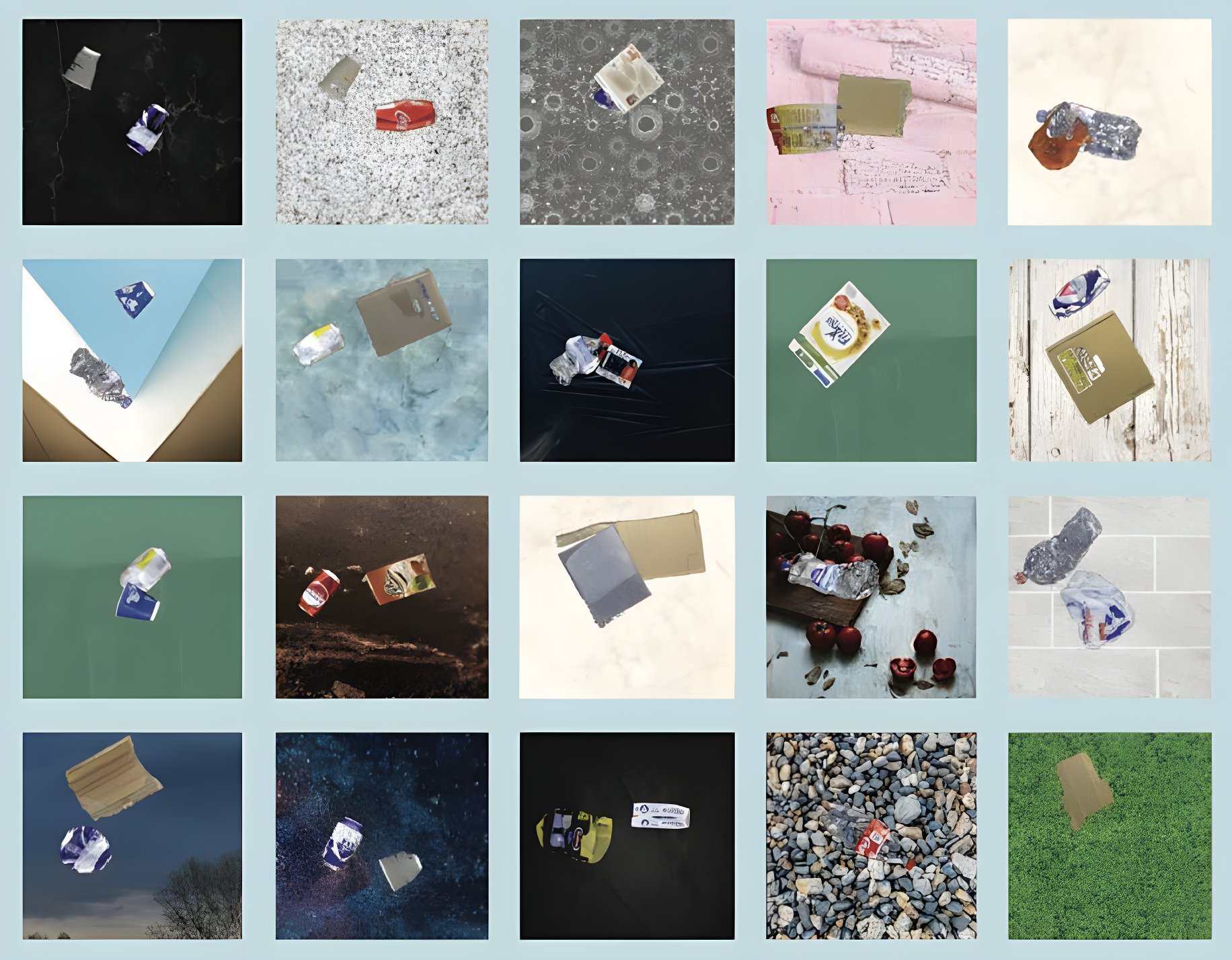}
    \caption{Dataset images with new colorful backgrounds  \cite{wasteSorting}}
    \label{fig:dataset}
\end{figure}

A key element of their research involves the development of a tailored open-source dataset, an indispensable resource for training our neural networks.
Notably, the available open-source datasets for public utilization, namely TrashNet \cite{trashnet}, and Taco \cite{taco}, are confined to outdoor waste classification scenarios. These datasets do not sufficiently address the inherent complexities of industrial demanding environments. On the contrary, the paper by Koskinopoulou \etal; embraces an inverse methodology geared toward automating the annotation process of addressing diverse problem entanglements.
Firstly, the mask that indicates each region of interest was assigned as an image annotation. In the second step, translation, rotation, and scaling were randomly applied to the image and the mask. Finally, pairs of objects from the previous stage, with their respective masks, were randomly chosen and positioned over new images with colorful backgrounds to develop complex problem instances (as shown in Figure \ref{fig:dataset}). 

\section{Design choices}


In this section, we will discuss our most significant intuitions and design preferences that allow us to attain our goals. 

\subsection{Models}
We present a concise overview of the main characteristics of the three neural networks employed in this work:
\begin{itemize}
  \item ENet (Efficient Neural Network): It is designed to perform semantic segmentation while being lightweight in terms of computing resources, with a size of $\sim$1.5MB and $\sim$360,000 parameters.  ENet utilizes innovative architectural features. Among them, spatial pooling is used to downsample feature maps efficiently. Moreover, the network's architecture consists of an initial convolutional layer, followed by several encoder and decoder blocks, with each block featuring bottleneck modules. Skip connections are employed to amalgamate feature maps from different layers, capturing low-level and high-level features.
  \cite{enet}
  \item BiSeNet V2 (Bilateral Segmentation Network): It is characterized by a two-branch network structure consisting of a detail branch and a semantic branch. It uses the Xception39 pre-trained model as a backbone. 
  The details branch is responsible for extracting spatial information, which is low-level information, while the semantic branch is designed to capture high-level semantics. Then, it is followed by an aggregation layer and a segmentation head. 
  This design strikes a balance between accuracy and efficiency. BiSeNet is known for its relatively compact model size (with a weight of $\sim$14MB and $\sim$5.2 million of parameters) compared to some other semantic segmentation networks. \cite{bisenet2}
  \item ICNet (Image Cascade Network): It is tailored for high-quality real-time semantic segmentation, even though it comes with a larger model size as a trade-off, with a dimension of $\sim$190MB and $\sim$48 millions of parameters. It employs a cascade of multiple branches to process input images at various scales, enabling it to capture both fine and coarse details in segmentation tasks. It incorporates a pyramid structure, efficiently capturing multi-scale features. Additionally, ICNet produces a hierarchical segmentation map with multiple resolution levels, which proves advantageous for delineating objects in various contexts. The cascade architecture enables multi-scale processing, allowing it to effectively handle objects of varying sizes. \cite{icnet}
\end{itemize}

\subsubsection{Custom ENet}
In the context of ENet, adjustments were made to the dimensions of specific layers in both the encoder and decoder. These changes were made possible by the model's size, which allowed for the addition and modification of layers without exceeding the 10MB limit. Specifically, we expanded the network's architecture adding a downsampling layer with the objective of achieving an output size of 256 x 32 x 32 for every bottleneck layer in stages 2 and 3, instead of the original configuration of 128 × 64 × 64. We will refer to this modified model as Custom ENet.


All the characteristics of the employed networks are summarised in Table \ref{tab:Net_summary} (FLOPS stands for floating point operations per second).

\begin{table}[ht] 
  \centering
  \begin{tabular}{c | c c c}  
    \toprule
   Net & Size (MB) &FLOPS & \# Parameters\\
    \midrule
    Enet & 1.45 & 11.88 & 363,132\\
    Custom Enet & 5.45 & 13.42 & 1,363,168\\
    BiSeNet & 13.38 & 74.89 & 5,188,485 \\
    ICNet & 189.96 & 99.35 & 47,489,184\\
    \bottomrule
  \end{tabular}
  \caption{Employed models main characteristics}
  \label{tab:Net_summary}
\end{table}

\subsection{Loss functions}
In semantic segmentation, a loss function is a critical component used to measure the dissimilarities between predicted pixel-wise class labels and ground truth labels in an image. 
Deep learning models are prone to bias towards the majority class. In the context of semantic segmentation, the background class (i.e. everything that is not part of the object of interest) often constitutes the majority class. This bias may lead the model to allocate excessive attention to the background, resulting in inaccurate segmentation of the foreground objects.
We implement three losses to address this problem: 
\begin{itemize}
  \item Focal loss: It's a function that introduces a modulation factor (regulated with a hyperparameter $\gamma$) into the Cross-entropy loss function, aimed at concentrating the learning process on challenging misclassifications. Essentially, it is a Cross-entropy loss that adapts its weighting dynamically. As the model gains more confidence in correctly classifying an example, this scaling factor progressively diminishes toward zero. In simpler terms, this factor automatically reduces the impact of straightforward examples during training and swiftly directs the model's attention to the more difficult instances.\cite{focal}
  \item Class-balanced focal loss: This is a re-weighting scheme based on the effective number of samples within each class to recalibrate the focal loss function.  
  The effective number reflects the idea that as the number of samples increases, the incremental benefit of a newly added image diminishes. It essentially represents the coverage of the data samples in representing the class in a feature space.\cite{class-balanced}
  \item Dice loss: Dice loss attaches similar importance to false positives and false negatives, dynamically adjusting weights to deemphasize easy-negative examples. \cite{dice}
\end{itemize}
Moreover, we utilize additional loss functions in order to enhance performance:
\begin{itemize}
    \item Lov{\'a}sz loss: Its primary objective is to promote the alignment of the predicted class with ground truth masks, leading to an improved mIoU value. \cite{lovasz}
    \item Focal-Lov{\'a}sz loss: Firstly introduced in \cite{focal+lovasz}, we aim to combine the benefits of both loss functions.
\end{itemize}

\subsection{Pruning and Quantisation}\label{section_pruning}
These two techniques are implemented to attain our objective of creating a lightweight model.

\subsubsection{Pruning}
Pruning is a neural network technique used to reduce the size and complexity of the network by selectively removing certain connections (weights) while trying to preserve the network's performance.
It works by creating a pruning mask for each layer of the model. The pruning mask determines which parameters should be pruned, so setting the values of the pruned connections or parameters to zero. \cite{pruning_pytorch,pruning_frankle2019lottery}
Pruning techniques can be categorized into two main types: unstructured (i.e. deleting individual parameters without their respective structure) and structured pruning (i.e. removing entire structures of parameters). We present an explanation of the methods employed:
\begin{itemize}
  \item Random Unstructured Pruning: During the pruning process, a certain percentage of weights are selected randomly and assigned a value of zero. This effectively eliminates those connections from the network, reducing its size and complexity.
  \item L1 Unstructured Pruning: It encourages sparsity in the neural network by zeroing out the weights tensors with the lowest L1-norm.
  \item Random Structured Pruning: It randomly removes entire neuron channels or filters from the neural network.
    \item Ln Structured Pruning: This pruning method, where n indicates the order of the norm, is an extension of L1 Pruning with a structured method. It encourages entire units to become exactly zero, according to their norms.
\end{itemize}

\subsubsection{Quantisation}
Quantization is a technique for performing computations and storing tensors at lower bit-widths than floating point precision. A quantized model executes some or all of the operations on tensors with reduced precision rather than full precision (floating point) values. $N$ float32 values are mapped to lower precision numbers that can be represented by $N$ uint8 values and 2 more single values: a float32 scale factor and an int32 zero\_point value. In this way, a float32 number can be reconstructed as $$val_{float32} = (val_{uint8} - zero\_point) \times scale$$ This process can lead up to a 4x reduction in memory size. \cite{quantisation_gholami2021survey}

Quantization can be classified into Dynamic and Static.
Dynamic quantization adapts the zero\_point and scale factors on the fly, making it powerful but computationally expensive. Static quantization, on the other hand, has fixed parameters for every input. Among the static quantization approaches, we examine two of them:
\begin{itemize}
    \item Post-Training Quantisation: It involves quantizing a pre-trained model after it has already been trained at high precision. This approach is simpler but may require fine-tuning to recover some performance lost due to quantization. 
    \item Quantisation-Aware Training: It minimizes the impact of quantization on model accuracy. During training, the model is exposed to quantized values, and gradients are computed accordingly. This helps the model adapt to the lower-precision representation while maintaining performance. 
\end{itemize}
In our work, we employ PTQ (Post-Training Quantisation) because it enables us to achieve a satisfactory reduction in model size.


\section{Experimental Results}

In this section, we present and analyze the experimental results obtained during the evaluation of our previously proposed methods. 
In each of our experimental setup, we employed the hyperparameters as detailed in the Table \ref{tab:hyperparameters}.
We will only showcase the ``optimal" results, defined by the criteria of achieving a high mIoU value while preserving minimal variance in the final epochs; this ensures the model's robustness towards the end of the training process.
Throughout the training, certain transformations such as scaling, random cropping, and random horizontal flips (with a 0.5 probability of occurring) are applied to the training dataset. 
Unless otherwise specified, the neural networks were trained for 100 epochs with the Cross-entropy loss.

\begin{table}[ht] 
  \centering
  \begin{tabular}{c c}  
    \toprule
        Parameters & Values\\
    \midrule 
     Learning rate (LR) & \{$5e^{-4},\, 5e^{-5},\, 5e^{-6}$\} \\
     Learning Rate decay ($LR_{D}$) & \{$None, 0.995$\} \\
     Step LR decay ($Step_{LR}$) & \{$None,\, 20,\, 25,\, 30,\, 50$\} \\ 
    \bottomrule
  \end{tabular}
  \caption{Hyperparameters used in training}
  \label{tab:hyperparameters}
\end{table}

\subsection{Binary segmentation}

Our first step was to train the segmentation networks to give a pixel-wise prediction for the waste area, specifically distinguishing between background and objects.

The optimal hyperparameter configurations have been outlined in Table \ref{tab:bin_seg_results}. 
As expected, BiSeNet and ICNet outperform Enet, primarily due to disparities in model complexity and size.

\begin{table}[ht]
    \centering
    \begin{tabular}{c|c c c|c}
        \toprule
         Net & LR & $LR_{D}$ & $Step_{LR}$ & mIoU\\
        \midrule
         Enet & $5e^{-6}$ & None & None & \textbf{0.780} \\
        BiSeNet & $5e^{-6}$ & None & None & \textbf{0.848} \\
        ICNet & $5e^{-5}$ & None & None & \textbf{0.830}\\
        \bottomrule
    \end{tabular}
    \caption{Binary segmentation experimental results}
    \label{tab:bin_seg_results}
\end{table}


\subsection{Instance segmentation}
Differently from the previous phase, our objective here is to categorize the waste materials into one of four possible labels: aluminum, paper, bottle, or nylon. 
As shown in Table \ref{tab:inst_seg_results}, compared to binary segmentation, there is a decrease in the mIoU as we now need to identify both objects against the background and classify them into material classes. It's worth noting that Aluminum is consistently recognized.
Regarding the nets, the performance of ENet and BiSeNet is comparable, while ICNet is the top-performing model.

\begin{figure}
    \centering
    \includegraphics[width=0.53\textwidth]{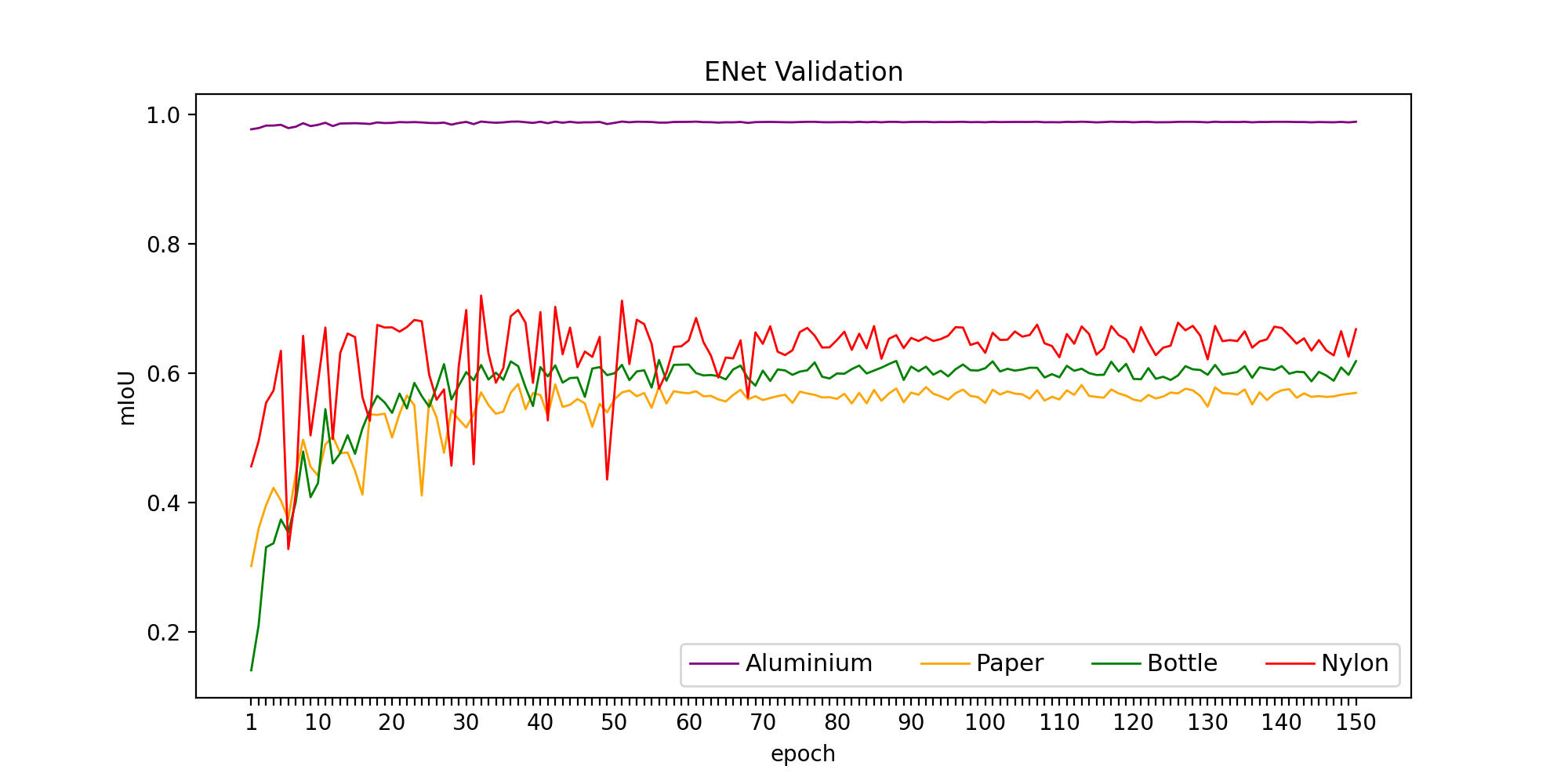}
    \caption{ENet validation: mIoU over epochs}
    \label{fig:plot_enet}
\end{figure}

\begin{figure}
    \centering
    \includegraphics[width=0.53\textwidth]{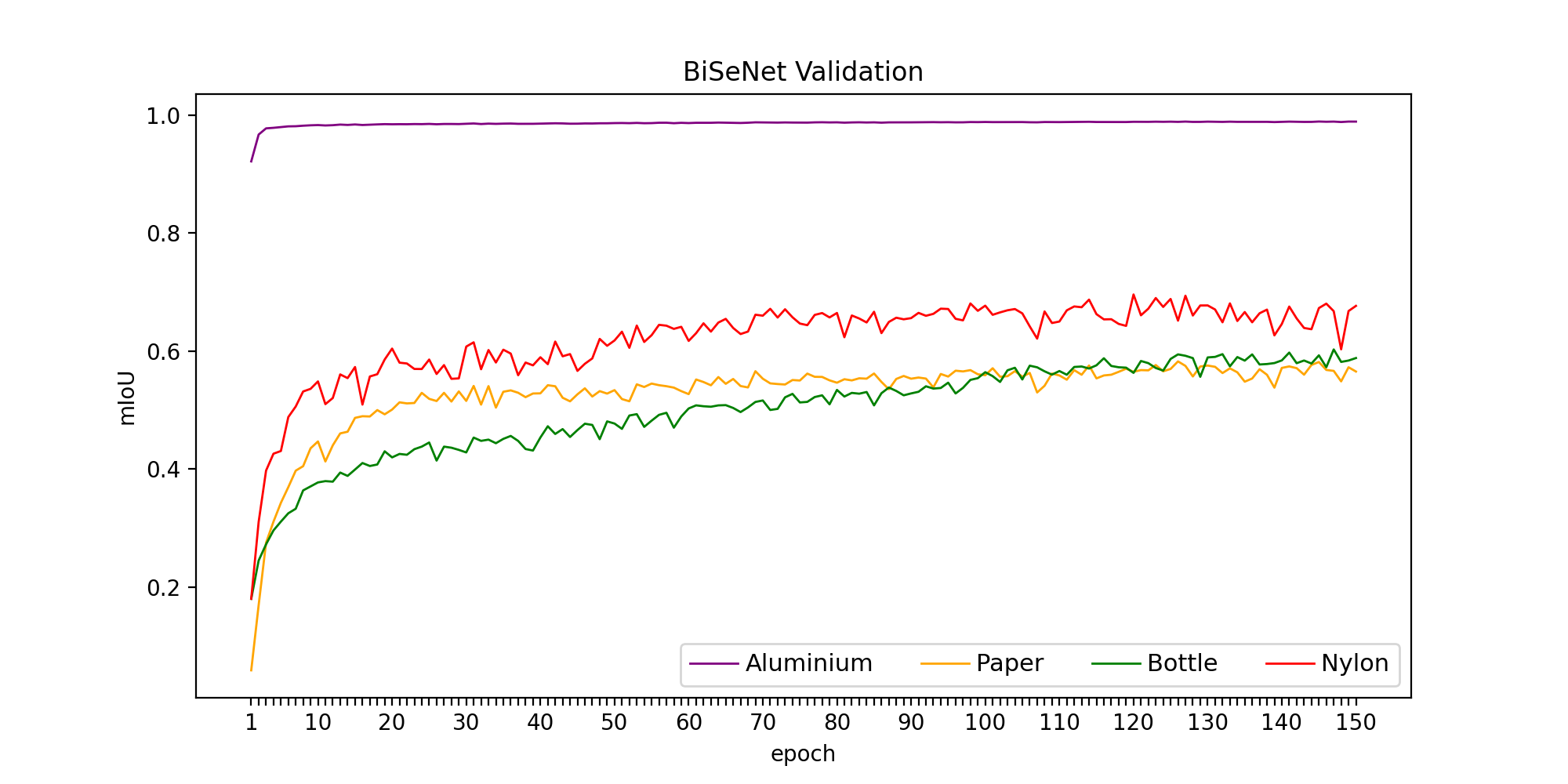}
    \caption{BiSeNet validation: mIoU over epochs}
    \label{fig:plot_bisenet}
\end{figure}

\begin{figure}
    \centering
    \includegraphics[width=0.53\textwidth]{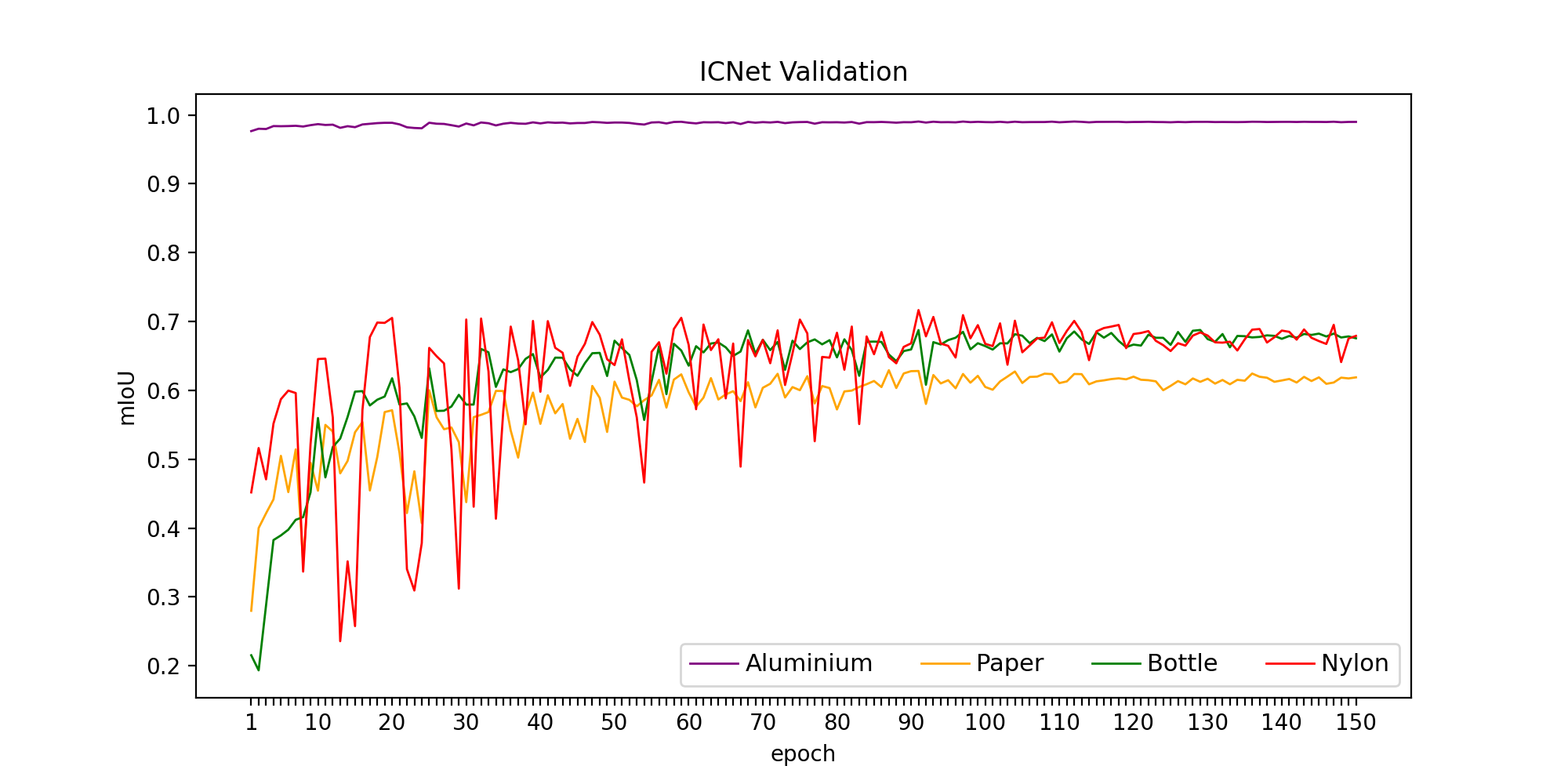}
    \caption{ICNet validation: mIoU over epochs}
    \label{fig:plot_icnet}
\end{figure}

Alongside Table \ref{tab:inst_seg_results}, we provide an analysis of the experimental training process in Figures \ref{fig:plot_enet},\ref{fig:plot_bisenet} and \ref{fig:plot_icnet}. 
A first consideration is that Aluminium, recognized since the first epochs, has demonstrated remarkable stability over the training process; this behavior can be explained by the repetitive shape of aluminum objects (mainly cans) and the reflectivity property of the material. 
At the outset, the recognition of the Bottle class faces initial challenges when compared to the other classes.
In each scenario, Paper consistently exhibits as the least performing.
Nylon, on the other hand, exhibits higher variance in performance, but interestingly, it emerges as the second-best performer in our analysis.
As we observe the training process, we notice that there is an equilibrium point reached after a certain number of epochs, i.e. model converges to the optimal result.
A remarkable observation is that BiSeNet, when lower learning rates are used, slowly converges to the final mIoU value without significant fluctuations.



\begin{table*}[ht]
    \centering
    \begin{tabular}{c|c c c c|c c c c |c}
        \toprule
         Net & LR & $LR_{D}$ & $Step_{LR}$ & \# Epochs & Aluminium & Paper & Bottle & Nylon & mIoU\\
        \midrule
        Enet & $5e^{-4}$ & 0.995 & 25 & 150 & 0.988 & 0.569 & 0.618 & 0.668 & \textbf{0.682} \\
        BiSeNet & $5e^{-6}$ & None & None & 150 &  0.989 & 0.565 & 0.588 & 0.676 & \textbf{0.679} \\
        ICNet & $5e^{-4}$ & 0.995 & 50 & 150 & 0.99 & 0.619 & 0.676 & 0.679 & \textbf{0.733}\\
        \bottomrule
    \end{tabular}
    \caption{Instance segmentation experimental results}
    \label{tab:inst_seg_results}
\end{table*}

\subsubsection{Custom ENet}

Table \ref{tab:customEnet} presents the results of our custom ENet implementation. It's noteworthy that despite being trained for a shorter duration, it converges to results that are on par with the standard ENet. However, under the same configuration (1), it fails to surpass the performance of the standard one. Additionally, even in the best-case scenario (2), the results do not meet expectations.

\begin{table*}[ht]
    \centering
    \begin{tabular}{c|c c c c|c c c c |c}
        \toprule
         Net & LR & $LR_{D}$ & $Step_{LR}$ & \# Epochs & Aluminium & Paper & Bottle & Nylon & mIoU\\
        \midrule
        Enet & $5e^{-4}$ & 0.995 & 25 & 150 & 0.988 & 0.569 & 0.618 & 0.668 & \textbf{0.682} \\
        Custom ENet (1)& $5e^{-4}$ & 0.995 & 25 & 100 & 0.987 & 0.548 & 0.586 & 0.653 & 0.667 \\
        Custom ENet (2)& $5e^{-4}$ & None & None & 100 & 0.986 & 0.56 & 0.603 & 0.628 & \textbf{0.679} \\
        
        \bottomrule
    \end{tabular}
    \caption{Instance segmentation experimental results of ENet and Custom ENet}
    \label{tab:customEnet}
\end{table*}

\subsubsection{Data Augmentation}

In this particular scenario, we aim to introduce additional transformations during the training process: random vertical flips (RVF) and color jittering (RCJ, i.e. light changes in brightness, contrast, saturation, and hue of the image). 
These alterations were applied respectively with probability 0.5 and 0.25. 

In Table \ref{tab:data_aug_results}, we present the results for the three models employing combinations of the aforementioned transformations (referred to as AT for brevity). 
Overall, the extra entation proved to be unsatisfactory, yielding outcomes that were inferior to those of the previous models.

\begin{table*}
    \centering
    \begin{tabular}{c |c| c c c |c c c c | c}
        \toprule
         Net & AT & LR & $LR_{D}$ & $Step_{LR}$ & Aluminium & Paper & Bottle & Nylon & mIoU\\
        \midrule
        Enet & RVF + RCJ & $5e^{-4}$ & None & None &  0.983 & 0.454 & 0.433 & 0.588 & 0.569 \\
        BiSeNet & RVF + RCJ & $5e^{-6}$ & None & None & 0.983 & 0.445 & 0.406 & 0.580 & 0.556 \\
        ICNet & RVF + RCJ & $5e^{-5}$ & None & None & 0.982 & 0.433 & 0.340 & 0.556 & 0.540\\
        \midrule
        Enet &  \textbf{RVF} & $5e^{-4}$ & 0.995 & 30 & 0.983 & 0.484 & 0.456 & 0.533 & \textbf{0.583} \\
        BiSeNet &  \textbf{RVF} & $5e^{-6}$ & 0.995 & 30 &  0.983 & 0.459 & 0.419 & 0.598 & \textbf{0.574} \\
        ICNet &  \textbf{RVF} & $5e^{-4}$ &0.995 & 30 & 0.982 & 0.490 & 0.414 & 0.555 & \textbf{0.567}\\
        \bottomrule
    \end{tabular}
    \caption{Data Augmentation experimental results}
    \label{tab:data_aug_results}
\end{table*}

\subsection{Loss Functions}

In the following section, we explore the use of various Loss Functions, a critical aspect of our experimental framework. We will present the results of our experiments, shedding light on the impact of these loss functions on the models' performance, when changing from the standard one (the Cross-entropy) that was employed in all previous experiments.
Notice that all the results reported for the Focal loss refer to an implementation with hyperparameter $\gamma = 2$ (we also tried 1 and 5, but 2 was the best one).

Table \ref{tab:losses_results} presents the results for each network trained with different loss functions. Notably, ENet consistently outperforms the others, with a specific mention going to the Class-balanced Focal and the Focal-Lov{\'a}sz losses, which yield a mIoU increase of 4.11\% and 7.48\%, respectively.

BiSeNet also demonstrates improved performance with certain loss functions when compared to Cross-entropy. Remarkably, the Lov{\'a}sz and Dice losses stand out, leading to improvements of up to nearly 6\%.

In contrast, ICNet experiences a notable decline in performance.

\begin{table*}
    \centering
    \begin{tabular}{c | c | c c c | c c c c | c}
        \toprule
         Net & Loss & LR & $LR_{D}$ & $Step_{LR}$ & Aluminium & Paper & Bottle & Nylon & mIoU\\
        \midrule
        Enet & Focal & $5e^{-4}$ & 0.995 & 25 & 0.987 & 0.612 & 0.653 & 0.524 & 0.680 \\
        BiSeNet & Focal & $5e^{-6}$ & None & None & 0.987  & 0.557 & 0.486 & 0.682 & 0.642 \\
        ICNet & Focal & $5e^{-4}$ & 0.995 & 50 & 0.988 & 0.441 & 0.616 & 0.666 & 0.663\\
        \midrule
        Enet & Lov{\'a}sz & $5e^{-4}$ & 0.995 & 25 & 0.988 & 0.605 & 0.629 & 0.656 & 0.689 \\
        BiSeNet & Lov{\'a}sz & $5e^{-6}$ & None & None &  0.989 & 0.591 & 0.618 & 0.689 & 0.702 \\  
        ICNet & Lov{\'a}sz & $5e^{-4}$ & 0.995 & 50 & 0.986 & 0.607 & 0.634 & 0.544 & 0.676\\
        \midrule
        Enet & Dice & $5e^{-4}$ & 0.995 & 25 & 0.984 & 0.546 & 0.514 & 0.535 & 0.600 \\
        \textbf{BiSeNet} & \textbf{Dice} & $5e^{-6}$ & None & None & 0.989  & 0.599 & 0.632 & 0.697 & \textbf{0.718} \\ 
        ICNet & Dice & $5e^{-4}$ & 0.995 & 50 & 0.987 & 0.604 & 0.644 & 0.606 & 0.686\\
        \midrule
        Enet & CBFL & $5e^{-4}$ & 0.995 & 25 & 0.989 & 0.598 & 0.653 & 0.682 & 0.710 \\
        BiSeNet & CBFL & $5e^{-6}$ & None & None & 0.988  & 0.542 & 0.546 & 0.613 & 0.642 \\
        ICNet & CBFL & $5e^{-4}$ & 0.995 & 50 & 0.985 & 0.349 & 0.621 & 0.710 & 0.662\\
        \midrule
        \textbf{Enet} & \textbf{Focal-Lov{\'a}sz} & $5e^{-4}$ & 0.995 & 25 & 0.99 & 0.60 & 0.682 & 0.734 & \textbf{0.733} \\
        BiSeNet & Focal-Lov{\'a}sz & $5e^{-6}$ & None & None &  0.989 & 0.582 & 0.584 & 0.67 & 0.685 \\
        \textbf{ICNet} & \textbf{Focal-Lov{\'a}sz} & $5e^{-4}$ & 0.995 & 50 & 0.988 & 0.532 & 0.663 & 0.638 & \textbf{0.693}\\
        \bottomrule
    \end{tabular}
    \caption{Loss functions experimental results}
    \label{tab:losses_results}
\end{table*}

\subsection{Pruning and Quantisation}

In the pursuit of reducing the size of the models, we explore the employment of pruning and quantization techniques, keeping in mind the strict memory constraint of 10MB.
It's worth noting that ENet, being inherently lightweight in terms of computational resources ($\sim$1.5MB), does not necessitate size reduction. Conversely, given that BiSeNet has an approximate size of 14MB, applying quantization or pruning alone suffices. 
Finally, given that ICNet is highly computationally demanding, with an estimated size of around 190MB, it requires both pruning and quantization to meet the specified constraints. 

The two models selected for pruning and quantization are the ones with the highest mIoU values shown in Tables \ref{tab:inst_seg_results} and \ref{tab:losses_results}. 

\subsubsection{Pruning} 

We tried the aforementioned methods (Paragraph \ref{section_pruning}) and it came out that only the L1 unstructured achieves acceptable results.
The implementation of pruning in PyTorch includes the hyperparameter "amount" that refers to the proportion of the network's parameters to be pruned, leading to an equal reduction in the model size. 
The outcomes of these experiments are depicted in Table \ref{tab:prun_results}.

Regarding the ICNet, it's important to note that in order to respect the 10MB constraint, it is necessary to use a pruning amount of at least 0.95 (corresponding to a size of 9.73MB). However, this adjustment results in a substantial degradation in performance, with the highest achievable mIoU standing at only 0.196 under these conditions.
Conversely, employing a low value for amount results in a favorable trade-off between parameter size and the final mIoU metric, albeit not adhering to the memory constraint.

It's important to note that BiSeNet not only respects  the 10MB limitation but also demonstrates that even though the higher amount value, it has a smaller decrease in mIoU compared to ICNet.

\begin{table*}
    \centering
    \begin{tabular}{c| c |c c c c | c c | c}
        \toprule
         Net & Amount & Aluminium & Paper & Bottle & Nylon & mIoU & Size (MB) & $\Delta$mIoU \% \\
        \midrule
        BiSeNet  & \textbf{0.3} & 0.988 & 0.566 & 0.614 & 0.675 & \textbf{0.688} & \textbf{9.38} & \textbf{-4.18\%}  \\
        \midrule
        ICNet  & 0.95 & 0.967 & 0.0 & 0.0 & 0.0 & 0.196 & 9.73 & -73.26\% \\
        ICNet  & \textbf{0.2} & 0.988 & 0.585 & 0.604 & 0.622 & \textbf{0.690} & \textbf{152.1} & \textbf{-5.91\%} \\
        ICNet  & \textbf{0.15} & 0.989 & 0.604 & 0.647 & 0.679 & \textbf{0.718} & \textbf{161.48} & \textbf{-2.04\% } \\
        \bottomrule
    \end{tabular}
    \caption{Pruning experimental results}
    \label{tab:prun_results}
\end{table*}


\subsubsection{Quantisation}

Post-training quantization was applied to both BiSeNet and ICNet.
At the outset, an attempt was made to employ Dynamic quantization, but it proved to be ineffective in the model. This is because PyTorch's implementation only affects Linear and Recurrent layers, which are absent in the ICNet architecture.
The results of these experiments are presented in Table \ref{tab:quantisation_result}.

BiSeNet demonstrates impressive outcomes when quantization is applied. The model size is reduced by over 75\%, and remarkably, this reduction has minimal impact on performance, as evidenced by a mere 1.53\% decrease in mIoU.

Quantization on ICNet proved to be highly effective, reducing its size by more than half with only a 1\% decrease in mIoU value.

\begin{table*}
    \centering
    \begin{tabular}{c|c c c c | c c | c c}
        \toprule
         Net & Aluminium & Paper & Bottle & Nylon & mIoU & Size (MB) & $\Delta$mIoU \% & $\Delta$Size \% \\
        \midrule
        BiSeNet  & 0.989 & 0.602 & 0.614 & 0.689 & \textbf{0.707} & \textbf{3.21} & \textbf{-1.53\%} & \textbf{-76.02\%} \\
        \midrule
        ICNet  & 0.990 & 0.609 & 0.653 & 0.697 & \textbf{0.725} & \textbf{90.69} & \textbf{-1.09\%}  & \textbf{-52.25\% } \\
        \bottomrule
    \end{tabular}
    \caption{Quantisation experimental results}
    \label{tab:quantisation_result}
\end{table*}

\subsubsection{Pruning + Quantisation}
As ICNet still did not meet the memory constraint, we attempted to combine Pruning and Quantisation: we applied quantization on the pruned model obtained with different values of amount (0.2, 0.4, and 0.6). Contrary to our initial expectations, we found that the combination of both techniques was less effective than applying Quantisation alone. The reduction in size was approximately the same, and the final mIoU value was slightly lower.
Our interpretation suggests that quantization, when applied after pruning, is less effective due to its impact on a reduced number of weights compared to the unpruned model.

\section{Conclusion}

The study's findings contribute to the development of efficient waste-sorting technology for improved recycling. 
The research emphasizes the importance of optimizing the model architecture and reducing memory footprint without compromising the segmentation accuracy required for effective waste sorting. The chosen network architectures successfully achieve accurate semantic segmentation for waste sorting, adhering to the stringent 10MB memory constraint in most cases.

Our findings reveal that ICNet excels under standard settings, outperforming the other models. Interestingly, when trained with different loss functions, both ENet and BiSeNet yield results that are comparable to ICNet.

entation and Custom ENet, meanwhile, didn't meet expectations.

A significant outcome of our research is the effect of Quantisation on BiSeNet and ICNet. This technique has made it possible to have an outstanding decrease in model size while maintaining the same performance.


{\small
\bibliographystyle{unsrt} 
\bibliography{egbib}
}



\end{document}